\documentclass[english,10pt,twocolumn,letterpaper]{IEEEtran}
\usepackage[T1]{fontenc}
\usepackage[latin9]{inputenc}
\usepackage{color}
\usepackage{array}
\usepackage{multirow}
\usepackage{amsmath}
\usepackage{graphicx}

\makeatletter

\providecommand{\tabularnewline}{\\}

\usepackage{times}
\usepackage{epsfig}
\usepackage{graphicx}
\usepackage{amsmath}
\usepackage{amssymb}

\usepackage[pagebackref=true,breaklinks=true,letterpaper=true,colorlinks,bookmarks=false]{hyperref}

\makeatother

\usepackage{babel}
\begin{document}
\title{Improving Structured Text Recognition with Regular Expression Biasing}
\author{Baoguang Shi, Wenfeng Cheng, Yijuan Lu, Cha Zhang, Dinei Florencio\\
Microsoft\\
\texttt{\small{}\{bashi, wenfeng.cheng, yijlu, chazhang, dinei\}@microsoft.com}}
\maketitle
\begin{abstract}
We study the problem of recognizing structured text, i.e. text that
follows certain formats, and propose to improve the recognition accuracy
of structured text by specifying regular expressions (regexes) for
biasing. A biased recognizer recognizes text that matches the specified
regexes with significantly improved accuracy, at the cost of a generally
small degradation on other text. The biasing is realized by modeling
regexes as a Weighted Finite-State Transducer (WFST) and injecting
it into the decoder via dynamic replacement. A single hyperparameter
controls the biasing strength. The method is useful for recognizing
text lines with known formats or containing words from a domain vocabulary.
Examples include driver license numbers, drug names in prescriptions,
etc. We demonstrate the efficacy of regex biasing on datasets of printed
and handwritten structured text and measures its side effects.
\end{abstract}

\section{Introduction}

A practical OCR system is often applied to recognize \emph{structured
text}, i.e. text that follows certain formats. Examples include dates,
currencies, phone numbers, addresses, etc. Structured text possesses
challenges to OCR recognizers, as such text contains digits and symbols
that are hard to separate, e.g. ``1'' (one), ``l'' (lower-case
L), and ``/'' (slash). Meanwhile, structured text possesses important
information and demands higher recognition accuracy.

\begin{figure}[t]
\begin{centering}
\includegraphics[width=0.85\linewidth]{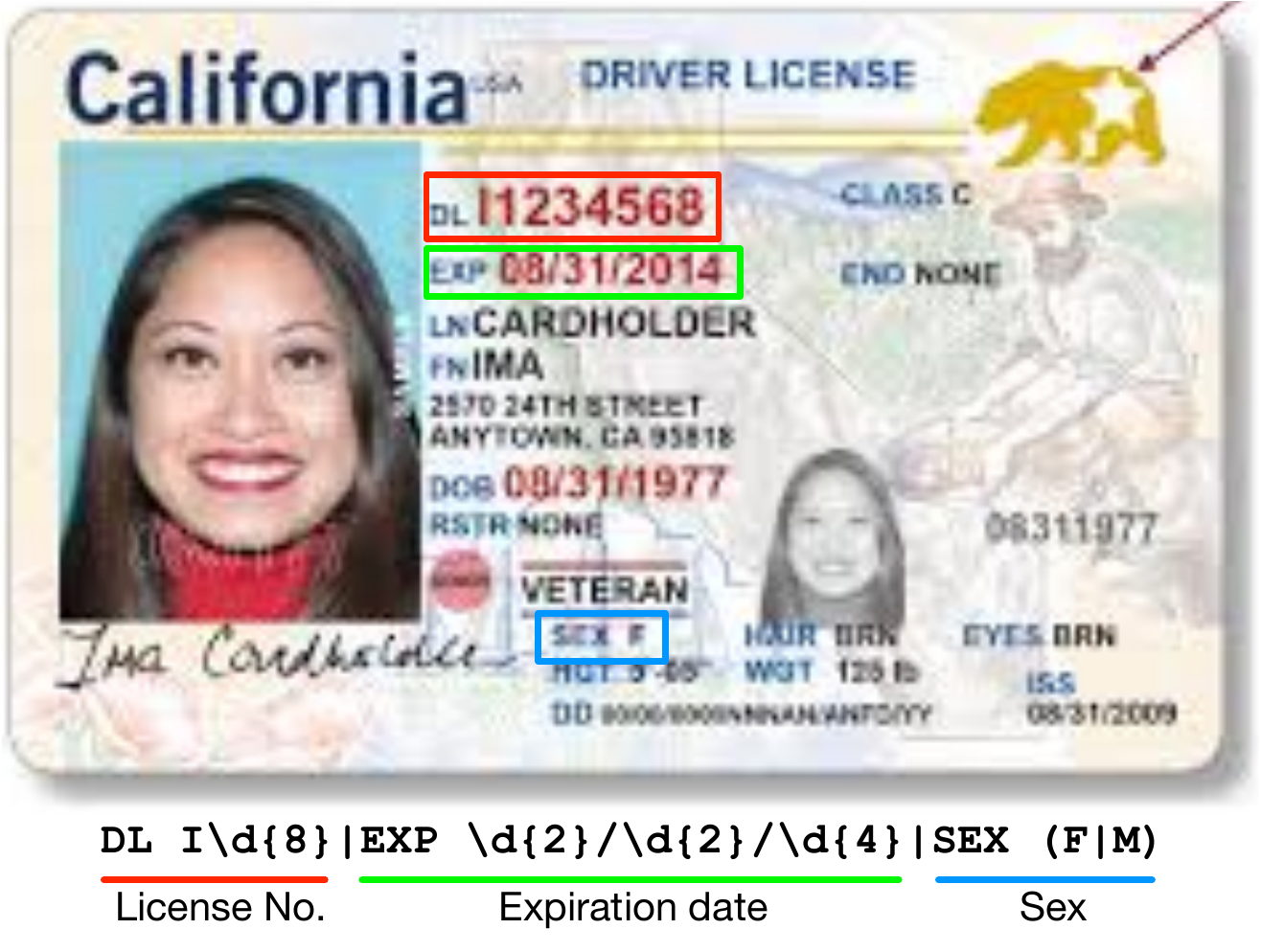}
\par\end{centering}
\caption{An illustration of regex biasing applied to recognizing a low-resolution
driver license scan. The specified regex consists of 3 parts, concatenated
by the OR (``|'') operator, biasing the recognition of license number,
expiration date, and sex fields. The fields impacted by the biasing
are highlighted by color-coded boxes. This image is a fake sample
from California DMV.}

\label{fig:illustration}
\end{figure}

\begin{figure*}[t]
\begin{centering}
\includegraphics[width=1\textwidth]{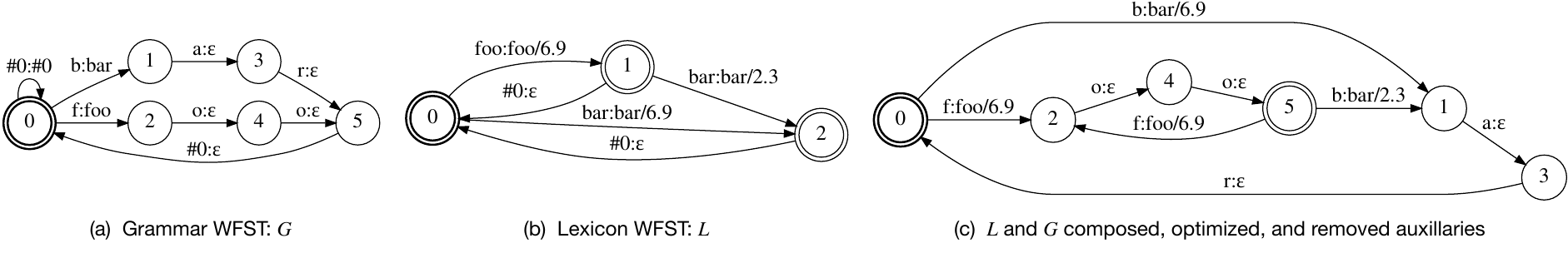}
\par\end{centering}
\caption{A mini WFST decoder representing the language model comprising words,
``foo'' and ``bar''. Thick-lined circles are start states; double-lined
circles are final states. Transition labels are formatted in ``<input
label>:<output label>/<weight>'', or ``<input label>:<output label>''
when weight is zero. The auxiliary symbol ``\#0'' is for disambiguation
when a word has more than one spelling (e.g. spelling in lower case
and upper case letters) \cite{povey2011kaldi}. In (a), the weight
values 6.9 is calculated from $-\log(0.001)$, meaning unigram probabilities
0.001 for words ``foo'' and ``bar''. The transition from state
1 to 2 means a 0.01 bigram probability for ``foo bar''.}
\label{fig:foobar}
\end{figure*}

Improving the recognition of structured text can be seen as a case
of domain adaptation. A common strategy for domain adaption is to
finetune recognition models with domain-specific data. However, finetuning
requires collecting a sufficiently large dataset in the same domain.
Therefore, finetuning can be very expensive, and impractical in many
cases due to the sensitivity of the data in the target domain.

In many cases, the format of the structured text is known in prior
and can be expressed by regular expressions. For example, California
car license plate number follows ``\texttt{\textbf{\footnotesize{}\textbackslash d{[}A-Z{]}\{3\}\textbackslash d\{3\}}}'',
meaning ``one digit, followed by three capital letters, then followed
by three digits''. The knowledge of the format may substantially
improve the recognition of structured text, as candidate characters
are limited by their positions and contexts.

In this paper, we propose to inject such knowledge into a text recognizer
by biasing it towards user-specified regexes. A biased recognizer
will favor text that matches the specified recognizer over other similar
candidates. Figure~\ref{fig:illustration} illustrates the biasing
using a regex describing the formats of license number, expiration
date, and sex. A recognizer biased as such will favor ``\texttt{\textbf{\footnotesize{}DL
}}\texttt{\textbf{\textcolor{red}{\footnotesize{}I}}}\texttt{\textbf{\footnotesize{}12345678}}''
over ``\texttt{\textbf{\footnotesize{}DL }}\texttt{\textbf{\textcolor{red}{\footnotesize{}1}}}\texttt{\textbf{\footnotesize{}12345678}}''
because the former matches the specified regex while the latter does
not. Consequently, a biased recognizer recognizes structured text
with significantly improved accuracy at the cost of a generally small
degradation on other text.

We realize regex biasing by expressing the regexes as a Weighted Finite-State
Transducer (WFST) \cite{MohriPR02} and use it to decode the outputs
of the recognition model. A single hyperparameter $\alpha\in\Re$
controls the weight of WFST and thus the biasing strength. Specifically,
a lower $\alpha$ makes the recognizer bias the specified regexes
with higher strength.

Regex biasing enables convenient domain adaption without training
data. Not only is this method cost-effective, but particularly useful
for situations where training data from the target domain is hard
to obtain due to its cost or sensitivity. We test the efficacy of
this method on both printed and handwritten text data. A side effect
of biasing is creating false positives, especially when biasing strength
is high. We measure this side effect in the experiments.

Since our method only concerns the WFST language model, it can be
applied to any recognizers that are compatible with WFST decoders.
We focus on the type of text recognizer that outputs a sequence of
character probabilities, e.g. \cite{ShiBY17,0001TDCSLH17}. Recognizers
of this kind have been shown to deliver state-of-the-art performance
on line-level recognition when working together with an n-gram language
model \cite{CongHHG19}.

The rest of the paper is organized as follows. In Section~\ref{sec:Related-works},
we discuss the related works. In Section~\ref{sec:Method}, we first
briefly present the basics of WFST, then introduce how to model regular
expressions as WFSTs and how to use them for decoding. In Section~\ref{sec:Experiments},
we present the experiment results, demonstrating the efficacy and
side-effects of regex biasing.

\section{Related works\label{sec:Related-works}}

Text recognition has attracted a lot of research effort in recent
years \cite{ShiBY17,ChengBXZPZ17,LiaoZWXLLYB19,ShiYWLYB19,AroraGWMSKCRBPE19,0001TDCSLH17,CaiH17,CongHHG19,LitmanATLMM20,QiaoZYZ020}.
These works model text recognition as a sequence prediction problem
and employ models such as convolutional network, LSTM, HMM (Hidden
Markov Model), and WFST for sequence modeling. Some works focus on
word-level scene text recognition \cite{ShiBY17,ChengBXZPZ17,LiaoZWXLLYB19,ShiYWLYB19}
while others address line-level handwriting recognition \cite{AroraGWMSKCRBPE19,0001TDCSLH17,CaiH17,CongHHG19}.
Our character recognition model is based on the CNN-LSTM-CTC framework
\cite{ShiBY17} and WFST for the language modeling part. It has a
similar architecture as the one in \cite{CaiH17}, which has been
demonstrated to outperform attention-based models when trained with
a large-scale dataset and coupled with a language model.

WFST has been highly successful in the research of speech recognition~\cite{MohriPR02}
and handwriting recognition~\cite{AroraGWMSKCRBPE19}. Because of
the flexibility of WFST, it can be used for modeling n-gram language
models, lexicons, etc. Our WFST building process follows the standard
recipe that involves lexicon and grammar modeling \cite{povey2011kaldi}.

The idea of biasing WFSTs with domain knowledge has been previously
explored in the speech recognition community \cite{AleksicAEKCM15,HaynorA20}.
We drew our inspiration from these works. \cite{AleksicAEKCM15} proposes
to improve the recognition of contact names by dynamic WFST replacement,
which we also used for injecting regex patterns. \cite{HaynorA20}
proposes to improve the recognition of numeric sequence by building
numeric grammar modeled by WFST, which is similar to our idea of improving
structured text with regex-defined grammars.

To the best of our knowledge, the idea of regex biasing has not been
previously proposed in the literature.

\section{Method\label{sec:Method}}

\begin{figure*}
\begin{centering}
\includegraphics[width=1\textwidth]{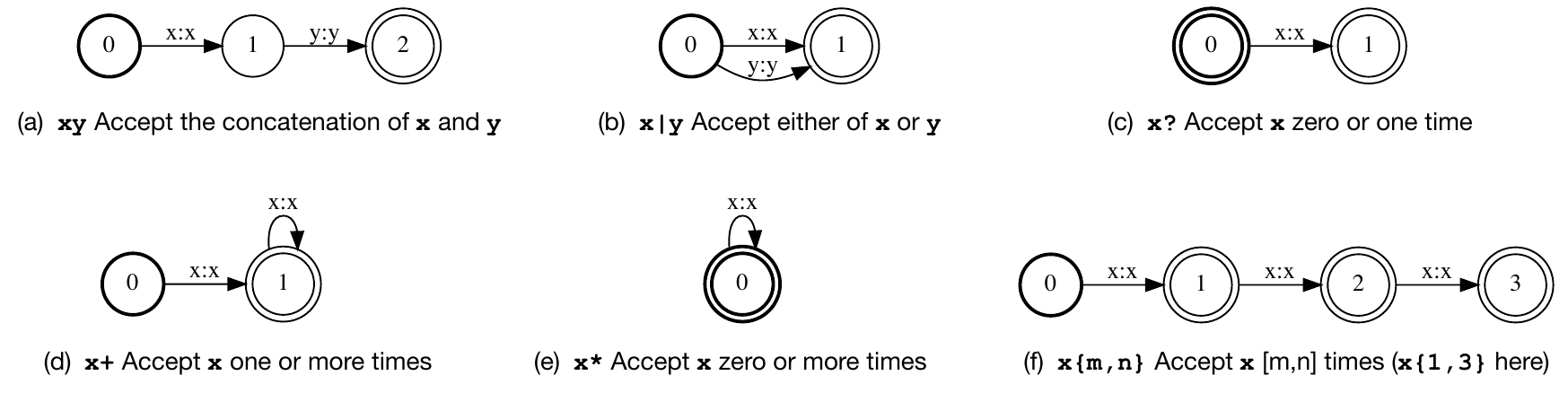}
\par\end{centering}
\caption{Example regex operations and their corresponding WFSTs. \texttt{\textbf{\footnotesize{}x}}
can be a symbol or a regular expression. In the examples above, \texttt{\textbf{\footnotesize{}x}}
is a symbol.}

\label{fig:regex-wfst}
\end{figure*}

\subsection{Background: WFST decoder}

WFST has a long-standing role in speech recognition \cite{MohriPR02}
for its language modeling capability. It provides a unified representation
of n-gram language model, lexicon, CTC collapsing rule, etc. We refer
the readers to \cite{MohriPR02} for a complete tutorial.

A WFST is a finite-state machine whose state transitions are labeled
with input symbols, output symbols, and weights. A state transition
consumes the input symbol, writes the output symbol, and accumulates
the weight. A special symbol $\epsilon$ means consuming no input
when used as an input symbol or outputting nothing when used as an
output label. Therefore, a path through the WFST maps an input string
to an output string with a total weight.

A common set of operations are available for WFSTs. \emph{Composition}
($\circ$) combines two WFSTs: Denoting the two WFSTs by $T_{1}$
and $T_{2}$, if the output space (symbol table) of $T_{1}$ matches
the input space of $T_{2}$, they can be combined by the composition
algorithm, as in $T=T_{1}\circ T_{2}$. Applying $T$ on any sequence
is equivalent to applying $T_{1}$ first, then $T_{2}$ on the output
of $T_{1}$. \emph{Determinization} and \emph{minimization} are two
standard WFST optimization operations. Determinization makes each
WFST state has at most one transition with any given input label and
eliminates all input $\epsilon$-labels. Minimization reduces the
number of states and transitions. The common optimization recipe combines
the two operations, as in $T'=\mathrm{optim}(T)=\mathrm{minimize}(\mathrm{determinize}(T))$,
and yields an equivalent WFST that is faster to decode and smaller
in size.

In the context of OCR, WFST can be used for decoding the output sequences
of a text recognition model. Below we demonstrate the WFST decoder
for a CTC-based \cite{GravesFGS06} recognizer. The WFST is composed
and optimized from three different WFSTs, denoted by $G$, $L$, and
$C$:
\begin{align}
T & =\mathrm{optim}(L\circ G)\label{eq:compose-clg}\\
T_{\mathrm{ctc}} & =\mathrm{optim}(C\circ T)\label{eq:make-ctc}
\end{align}
Here, $G$ (grammar) models n-gram probabilities. Its input and output
symbols are words (or sub-word units, such as BPE \cite{SennrichHB16a})
and its transition weights represent n-gram probabilities. $L$ models
lexicon, i.e. the spelling of every word in $G$. Its input space
is the set of characters supported by the text recognizer and its
output space is the words modeled by $G$. Since a CTC-based recognizer
outputs extra \emph{blank} symbols, an extra WFST $C$ is left-composed
to perform the ``collapsing rule'' of CTC. In practice, $C$ is
realized by inserting states and transitions that consume all blanks
and repeated characters to $L\circ G$. In Figure \ref{fig:foobar},
we illustrate $L$, $G$ and $T$ on a mini language model involving
only two words ``foo'' and ``bar''.

Decoding with WFST is to find the most probable word (or sub-word
unit) sequence $\hat{W}=w_{1},...,w_{M}$ given the character observations
$X=x_{1},...,x_{n}$ output by the CTC recognition model:

\begin{equation}
\hat{W}=\arg\min_{W}(\lambda\log p(X|W)+\log p(W))\label{eq:decode}
\end{equation}
Here, $\lambda$ (known as \emph{acoustic weight} in speech) controls
the weight of the character observations. The most probable path can
be approximated by the beam search algorithm. Open-source toolkits
such as Kaldi~\cite{povey2011kaldi} provide highly efficient decoding
implementations.

\subsection{Modeling regex as WFST}

Regular expressions are widely used in computer science for specifying
search patterns. A regex expression can be translated into a deterministic
finite automaton (DFA) by a regex processor, such as the Thompson's
construction algorithm \cite{AhoSU86}. Since WFST is also finite
automaton, we can convert the DFA of a regex into a WFST by turning
every transition label into a pair of identical input and output labels
and assign a unit weight. The resulting unweighted WFST is denoted
by $R$.

Figure~\ref{fig:regex-wfst} demonstrates some basic regex operations
and their corresponding WFSTs. Using these operators we can build
regex to match complex patterns. In practice, we rely on the open-source
grammar compiler \emph{Thrax} \cite{RoarkSARST12} to compile regexes
directly to WFSTs. The syntax of Thrax is similar to the common regex
syntaxes such as that in Python, but it also supports more advanced
features such as functions. The pseudo regex syntax we have used so
far can be easily translated into equivalent Thrax statements. For
example, the digit matcher ``\texttt{\textbf{\footnotesize{}\textbackslash d}}''
in Figure~\ref{fig:illustration} is replaced by a constant ``\texttt{\textbf{\footnotesize{}DIGIT}}''
in Thrax's syntax, which is defined as ``\texttt{\textbf{\footnotesize{}DIGIT
= ''0''|''1''|...|''9''}}''.

We make regex WFSTs have small or even negative transition weight
values so that the paths through regex WFSTs will be favored by the
decoder. In practice, we use a length-linear function to assign the
weights in the WFST transitions. This implemented by left-composing
a scoring WFST $S$ with the unweighted regex WFST $R$:

\[
T_{r}=S_{\alpha}\circ R
\]
Here, $S_{\alpha}$ is a scoring WFST that has a single state that
is both start and final state and connects a number of self-loop transitions
where the input and output labels are the supported symbols (characters).
The weights of these transitions are set to a constant $\alpha$.
After the composition, the total weight of a path in $T_{r}$ for
a matching text string will be $n\alpha$, where $n$ is the length
of the string. In this way, we can control the biasing strength by
adjusting $\alpha$: lowering $\alpha$ increases the biasing strength.

\subsection{Decoding with regex biasing}

The weighted regex WFST $T_{r}$ cannot be used directly for decoding
since it only accepts text matching the regex. We combine $T_{r}$
with the base language model $T$ so that the decoder can output any
text.

To achieve this, we first modify $T$ to add special transitions that
are labeled with a \emph{nonterminal} \cite{chomsky2002syntactic}
symbol \texttt{\textbf{\small{}\$REGEX}}. The modified WFST $T_{\mathrm{root}}$
is known as the \emph{class-based language model} \cite{BrownPdLM92}.
One way to add \texttt{\textbf{\small{}\$REGEX}} is to add it as a
unigram word to the grammar WFST $G$ so that regex biasing can be
applied to part of a sentence, while the rest of the sentence is scored
by $G$. The lexicon WFST $L$ is also modified to add the spelling
of \texttt{\textbf{\small{}\$REGEX}}, which is one epsilon (``$\epsilon$'').
The modified grammar and lexicon WFSTs are denoted by $G'$ and $L'$,
respectively. Then we have $T_{\mathrm{root}}=\mathrm{optim}(G'\circ L')$.

\begin{figure}
\begin{centering}
\includegraphics[width=1\columnwidth]{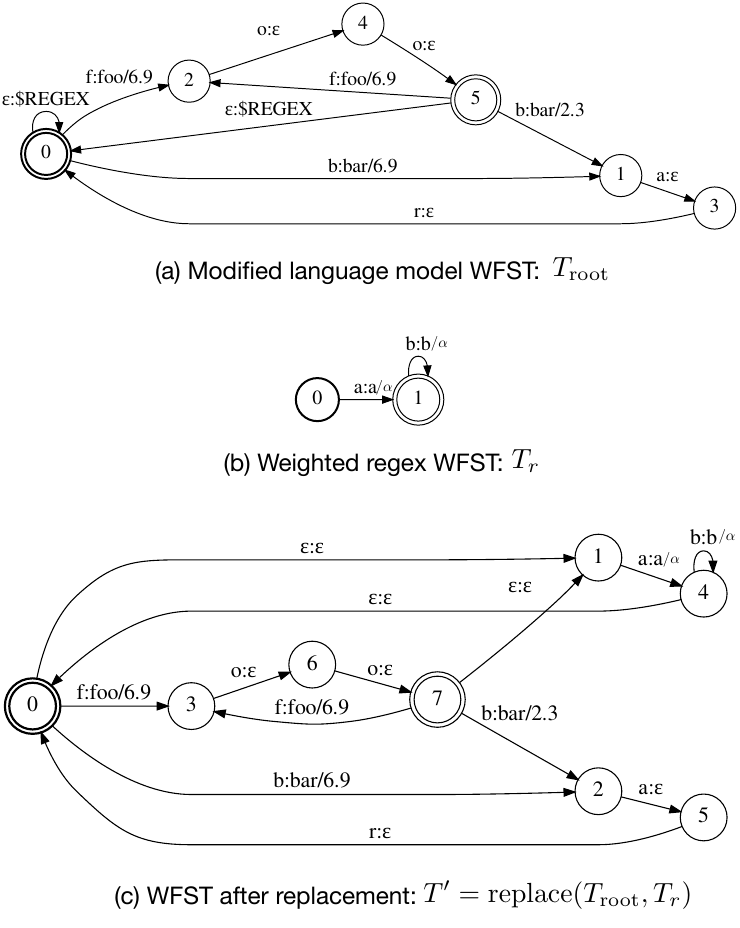}
\par\end{centering}
\caption{Illustration of WFST replacement. (a) Modified WFST $T_{\mathrm{root}}$
with nonterminal symbol \texttt{\textbf{\small{}\$REGEX}}; (b): Regex
WFST $T_{r}$ representing regex ``\texttt{\textbf{\small{}ab{*}}}''
with biasing strength $\alpha$; (c) Unoptimized WFST after replacement.}

\label{fig:replacement}
\end{figure}

$T_{\mathrm{root}}$ and $T_{r}$ can be combined using the WFST \emph{replacement}
operation:
\[
T'=\mathrm{replace}(T_{\mathrm{root}},T_{r})
\]
which replaces transitions labeled with \texttt{\textbf{\small{}\$REGEX}}
with its corresponding WFST $T_{r}$. Figure~\ref{fig:replacement}
illustrates this process on the aforementioned mini language model.
The modified base language model $T_{\mathrm{root}}$ has two additional
transitions with \texttt{\textbf{\small{}\$REGEX}}. After replacement,
state 0 and state 7 in $T'$ (corresponding to state 5 in $T_{\mathrm{root}}$)
both have a transition to state 1, effectively acting as the entry
and return points of $T_{r}$. After the replacement, we can make
$T'$ into a CTC-compatible decoder $T'_{\mathrm{ctc}}$ using the
composition in Equation~\ref{eq:make-ctc}.

In practice, we use an operation called \emph{dynamic replacement}
\cite{AllauzenRSSM07} to perform the combination. We first build
CTC-compatible WFSTs from $T_{\mathrm{root}}$ and $T_{r}$ respectively
(denoted by $T_{\mathrm{root}}^{c}$ and $T_{r}^{c}$ respectively).
Then, during decoding, transitions in $T_{\mathrm{root}}^{c}$ is
replaced by $T_{r}^{c}$ on-demand, as in $T'=\mathrm{dyreplace}(T_{\mathrm{root}}^{c},T_{r}^{c})$.
Since the main language model WFST $T_{\mathrm{root}}$ may contain
millions of states and transitions and is costly to update, dynamic
replacement allows us to fix the main language model and only update
the regex WFST.

\section{Experiments\label{sec:Experiments}}

In this section, we demonstrate the efficacy of regex biasing through
different combinations of dataset and regex setting. We focus on the
improvements on the structured text of interest as well as the side
effects, i.e. degradation on other text.

\subsection{Evaluation datasets}

Throughout the experiments, recognition is performed on text lines
rather than isolated words. This setting fits real-world scenarios
and is necessary for regex biasing since structured text often comes
in multiple words that are separated by space (e.g. ``Date of Birth:
Aug 1, 1990''). To the best of our knowledge, there are no public
datasets of such structured text. Therefore, we collect two datasets
of printed text and simulate structured text recognition on a handwritten
text dataset.

\textbf{Driver Licenses} This dataset consists of scans of US driver
licenses from different states. The licenses are fake samples we collected
from the Internet. To simulate the real-world imaging conditions,
we printed the samples on paper and took photos of them with phone
cameras. We ran an in-house text detector to extract text lines and
label them manually. There are 737 text lines from 50 distinct licenses.
Many lines contain structured text, such as date of birth and driver's
license number. To simulate poor imaging conditions, we further augmented
each image by degrading image quality and got 4422 text line images
for the final dataset.

\textbf{Passport MRZ} This dataset contains 8040 text lines extracted
from the machine-readable zones (MRZ) of a collection of passport
scans from different countries. Each MRZ contains two text lines.
Usually, MRZs are scanned by specialized passport readers and their
images are in high resolution. But in this dataset, images have much
lower quality in terms of resolution, lighting condition, etc.

\textbf{IAM} \cite{MartiB02} This dataset has been a standard dataset
for handwriting recognition. The test set of IAM contains 1861 handwritten
text lines. Some text lines of IAM contain out-of-vocabulary words
such as people's names. We use this dataset to test the efficacy of
biasing a word list.

\begin{table}
\caption{Regex definition for the driver license dataset in Thrax's syntax.
``UPPER'' and ``DIGIT'' are predefined constants, meaning upper-case
letters and digits respectively.}

\medskip{}

\includegraphics[width=1\columnwidth]{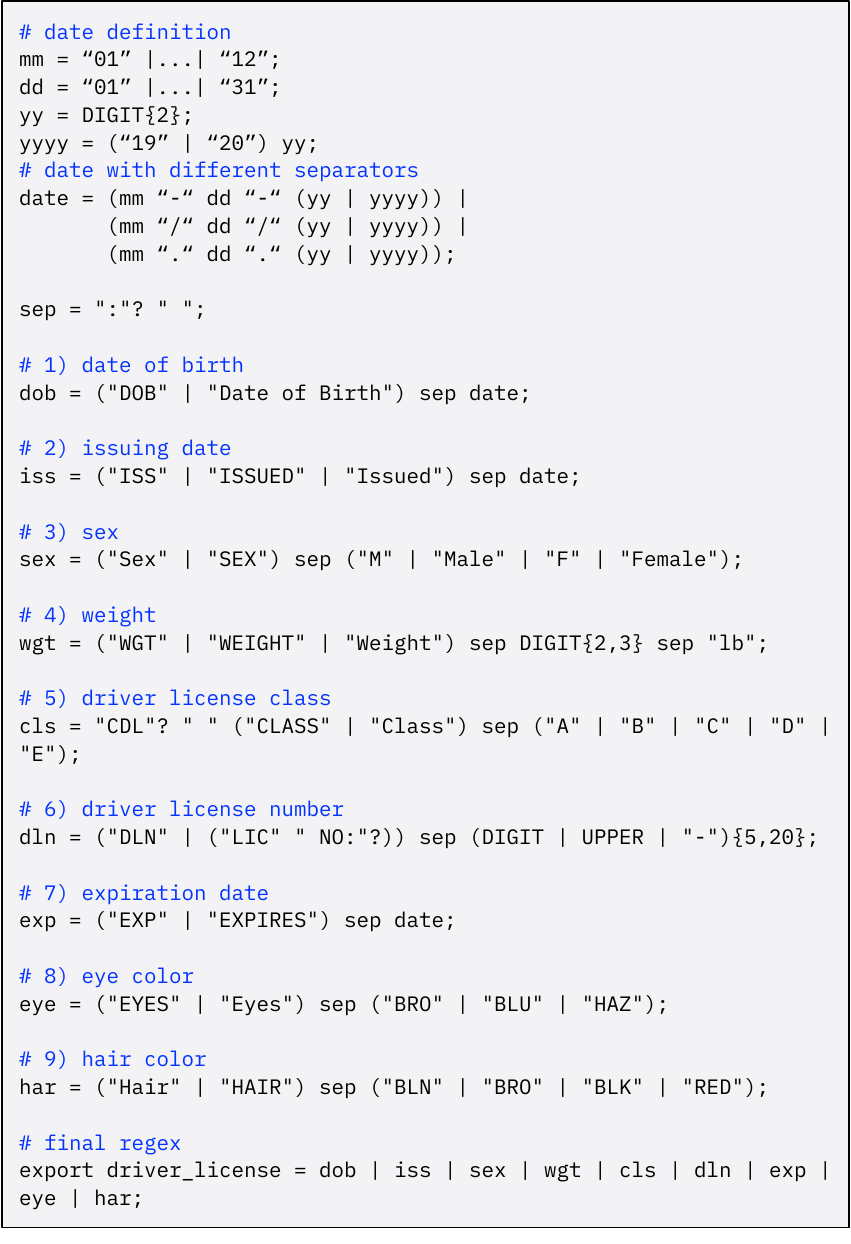}

\label{tab:code-dl}
\end{table}

\subsection{Recognition models}

Our recognizer consists of a character-level recognition model (character
model) and a language model. The character model follows the ConvNet-LSTM-CTC
design \cite{ShiBY17}. We use the compact model architecture from
\cite{0001TDCSLH17}. Since our model needs to be able to perform
line-level recognition, we trained our model on an internal dataset
of 1.6 million text lines. Images in this dataset come from multiple
sources, such as documents, receipts, scene images, and are in printed
style. This model is referred to as the \emph{printed model}. Similarly,
we train another model for handwriting recognition on a dataset consisting
of 248k handwritten text lines (\emph{handwriting model}).

We use the Kaldi toolkit~\cite{povey2011kaldi} to build the language
model. The language model is trained on a large text corpus comprising
67 million lines of text, collected from various sources. The lexicon
WFST is built from a vocabulary of 100k words and sub-word units,
learned from the training data using the byte-pair encoding algorithm~\cite{SennrichHB16a}
implemented in the sentencepiece library \cite{KudoR18}. The grammar
WFST is built from an n-gram model containing unigrams and bigrams.

We use the OpenGrm toolkit~\cite{RoarkSARST12} to build regex WFSTs
and convert them into CTC-compatible decoders using a Python binding
of Kaldi~\cite{pykaldi}. Building a regex WFST typically takes a
few seconds.

\subsection{Regex biasing for driver licenses}

The formats of US driver licenses differ from state to state. We set
regexes for a common set of fields, including date of birth, issuing
date, expiration date, etc. The full regex is in Table~\ref{tab:code-dl}.
We set different regexes 9 types of fields. The final regex is the
concatenation of the regexes using the OR operator.

We tested the WFST with different biasing strengths $\alpha$ and
measure their performance in terms of word error rate (WER). The recognition
model is the printed model. To test the efficacy of regex biasing
on matching text lines and its side effects on other text lines, we
divide the full set into two subsets, ``fields'' and ``non-fields'',
and calculate their WERs respectively.

\begin{table}
\caption{Word error rate (\%) on the driver license dataset. ``Full'' is
the full dataset; ``Fields'' is the subset where text matches specified
regex, partially or fully; ``Non-fields'' is the subset where the
text does not match the regex. Relative WER change to ``no bias''
WER is displayed in percentages.}

\medskip{}

\begin{centering}
\begin{tabular}{|c|c|c|c|c|c|c|}
\hline 
{\footnotesize{}Subset} & {\footnotesize{}No bias} & {\footnotesize{}$\alpha=0$} & {\footnotesize{}$-1$} & {\footnotesize{}$-3$} & \textbf{\footnotesize{}$-5$} & {\footnotesize{}$-10$}\tabularnewline
\hline 
\hline 
\multirow{2}{*}{{\footnotesize{}Full}} & \multirow{2}{*}{{\footnotesize{}5.5}} & {\footnotesize{}5.0} & {\footnotesize{}4.9} & {\footnotesize{}4.9} & {\footnotesize{}4.9} & {\footnotesize{}5.6}\tabularnewline
\cline{3-7} \cline{4-7} \cline{5-7} \cline{6-7} \cline{7-7} 
 &  & \textcolor{blue}{\footnotesize{}-9\%} & \textcolor{blue}{\footnotesize{}-11\%} & \textcolor{blue}{\footnotesize{}-11\%} & \textcolor{blue}{\footnotesize{}-11\%} & \textcolor{red}{\footnotesize{}+2\%}\tabularnewline
\hline 
\multirow{2}{*}{{\footnotesize{}Fields}} & \multirow{2}{*}{{\footnotesize{}5.7}} & {\footnotesize{}4.1} & {\footnotesize{}3.8} & {\footnotesize{}3.8} & {\footnotesize{}3.7} & {\footnotesize{}4.1}\tabularnewline
\cline{3-7} \cline{4-7} \cline{5-7} \cline{6-7} \cline{7-7} 
 &  & \textcolor{blue}{\footnotesize{}-28\%} & \textcolor{blue}{\footnotesize{}-33\%} & \textcolor{blue}{\footnotesize{}-33\%} & \textcolor{blue}{\footnotesize{}-35\%} & \textcolor{blue}{\footnotesize{}-28\%}\tabularnewline
\hline 
\multirow{2}{*}{{\footnotesize{}Non-fields}} & \multirow{2}{*}{{\footnotesize{}5.4}} & {\footnotesize{}5.4} & {\footnotesize{}5.4} & {\footnotesize{}5.4} & {\footnotesize{}5.5} & {\footnotesize{}6.4}\tabularnewline
\cline{3-7} \cline{4-7} \cline{5-7} \cline{6-7} \cline{7-7} 
 &  & {\footnotesize{}0\%} & {\footnotesize{}0\%} & {\footnotesize{}0\%} & \textcolor{red}{\footnotesize{}+2\%} & \textcolor{red}{\footnotesize{}+19\%}\tabularnewline
\hline 
\end{tabular}
\par\end{centering}
\label{tab:wer-dl}
\end{table}

Table~\ref{tab:wer-dl} summarizes the results. Without regex biasing,
the WER on the full set is 5.5, and the WERs on the two subsets are
close. When regex biasing is applied with $\alpha=0$, the WER on
fields drops to 4.1\%, while the WER on non-fields does not change.
It is worth noting that, since the weights in the main language model
are mostly positive, setting $\alpha=0$ still creates a bias.

As we further increase the biasing strength, the WER on fields drops
to 3.7\% when $\alpha=-5$, reducing 35\% of the errors compared with
no biasing, while the WER on non-fields slightly increases to 5.5\%
due to false positives, i.e. text that is mistakenly recognized to
match the regex. The overall WER is at its lowest 4.9\%. 

Beyond $\alpha=-5$, the number of false positives drastically increases,
and the overall WER increases, indicating that the regex biasing is
too strong and creates many false positives. On the other hand, the
WER on matching fields also increases beyond $\alpha=-5$. There are
two factors behind this increase: 1) False positives appear in the
part where the text does not match the regex; 2) Because beam search
sets a maximum weight difference between the allowed candidates and
the best one, a lower $\alpha$ will lead to the pruning of some correct
paths. As we increase the beam size, the WER on fields gets closer
to that of $\alpha=-5$.

The optimal setting of $\alpha$ depends on the application. Lower
$\alpha$ improves the accuracy of text matching the regex at the
cost of increased false positives. But as a rule of thumb, $\alpha$
in the range of $[-3,0]$ significantly reduces errors on matching
text and brings limited regression.

\begin{figure*}
\begin{centering}
\includegraphics[width=1\textwidth]{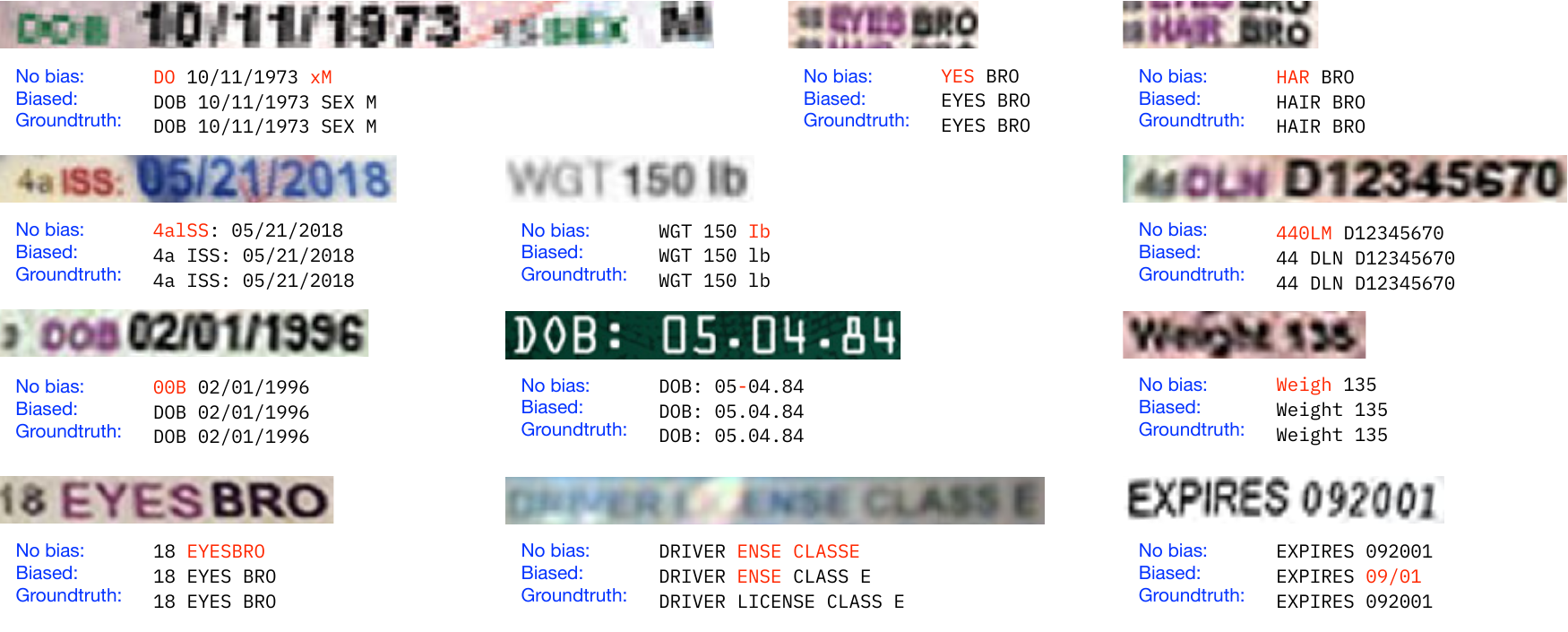}
\par\end{centering}
\caption{Examples of regex biasing on the driver license dataset. ``Biased''
results are from regex biasing with $\alpha=-5$. Recognition errors
are highlighted in red color.}

\label{fig:dl-examples}
\end{figure*}

Figure~\ref{fig:dl-examples} shows the efficacy and side effects
of regex biasing through examples. With regex biasing, the recognizer
correctly recognizes some highly challenging examples which are hard
to recognize without prior knowledge of their formats. On the other
hand, over biasing leads to more false positives. A failure case of
such is displayed in the last row.

\begin{table}
\caption{Regex definition for passport MRZ. Definitions for ``yy'', ``mm'',
and ``dd'' are reused from Table~\ref{tab:code-dl}.}

\medskip{}

\includegraphics[width=1\columnwidth]{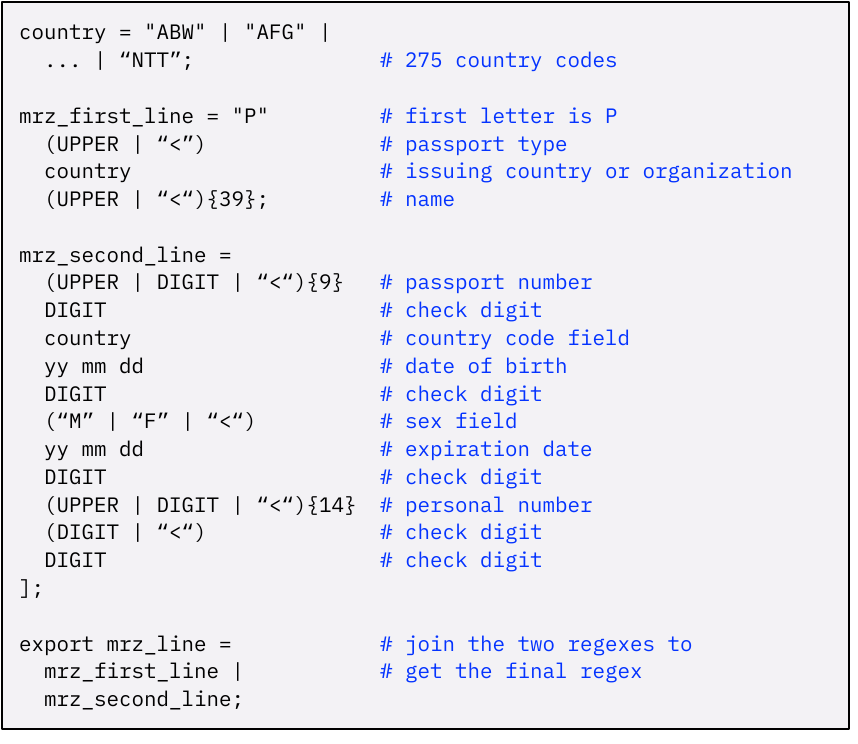}

\label{tab:code-mrz}
\end{table}

\subsection{Regex biasing for passport MRZ}

Passport MRZ follows the international standard ISO/IEC 7501-1. The
standard specifies the number of characters (44) and the allowed characters
at each position. Using the standard, we wrote the regexes shown in
Table~\ref{tab:code-mrz} and compile the final regex \texttt{\textbf{\small{}mrz\_line}}
into a WFST.

We measure the recognition performance on this dataset using character
error rate (CER). We used the printed model for this dataset. Table~\ref{tab:cer-mrz}
summarizes the results under different $\alpha$. Again, we observe
a significant improvement in character error rate after biasing. Character
error rates are reduced by 36\% at $\alpha=-2$. We also observed
that the biased recognizer avoided many common mistakes in the unbiased
recognizer, such as confusing ``0'' with ``O''. As $\alpha$ further
decreases, we see an increase in CER, also caused by early pruning
in the beam search.

\begin{table}
\caption{Character error rate (\%) on the MRZ dataset.}

\medskip{}

\begin{centering}
\begin{tabular}{|c||c|c|c|c|c|c|}
\hline 
 & {\footnotesize{}no bias} & {\footnotesize{}$\alpha=0$} & {\footnotesize{}$-1$} & {\footnotesize{}$-2$} & {\footnotesize{}$-3$} & {\footnotesize{}$-5$}\tabularnewline
\hline 
\hline 
\multirow{2}{*}{{\footnotesize{}Full}} & \multirow{2}{*}{{\footnotesize{}8.5}} & {\footnotesize{}5.4} & {\footnotesize{}5.4} & {\footnotesize{}5.4} & {\footnotesize{}5.5} & {\footnotesize{}6.8}\tabularnewline
\cline{3-7} \cline{4-7} \cline{5-7} \cline{6-7} \cline{7-7} 
 &  & \textcolor{blue}{\footnotesize{}-36\%} & \textcolor{blue}{\footnotesize{}-37\%} & \textcolor{blue}{\footnotesize{}-36\%} & \textcolor{blue}{\footnotesize{}-35\%} & \textcolor{blue}{\footnotesize{}-20\%}\tabularnewline
\hline 
\end{tabular}
\par\end{centering}
\label{tab:cer-mrz}
\end{table}

\subsection{Biasing a domain-specific vocabulary}

Regex biasing can be used for biasing a domain-specific vocabulary.
A list of words can be expressed as a regex by joining them with the
OR operator, as in ``\texttt{\textbf{\footnotesize{}<word1>|<word2>|
... |<wordN>}}''. Biasing to such regexes is useful for recognizing
words in a domain-specific vocabulary. For example, when trying to
recognize handwritten prescriptions, we can set a domain vocabulary
of common drug names and bias them by setting $\alpha$.

We use the IAM dataset to simulate such a scenario. Part of the text
in IAM comes from a novel. Some character names have uncommon spellings
and therefore are recognized with lower accuracies. Using the 37 character
names, we set the regex shown in Table~\ref{tab:code-iam}.

\begin{table}
\caption{Regex definition for IAM.}

\medskip{}

\includegraphics{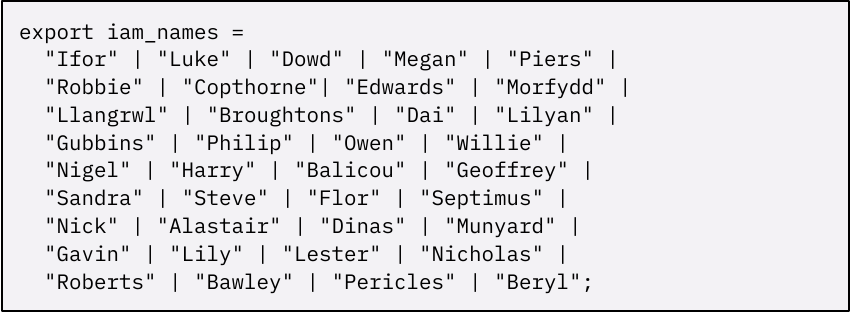}

\label{tab:code-iam}
\end{table}

We measure the performance of regex biasing using two metrics: 1)
the error rate on the 38 names, calculated by the sum of insertion,
deletion, and substitution error, divided by the number of name appearances;
2) the WER on the whole dataset. The results are summarized in Table~\ref{tab:results-iam}.
As the biasing strength increases, we see a significant drop in the
errors on the names. The WER on the full set also goes down until
$\alpha=-5$. As $\alpha$ further decreases, the errors on names
keep decreasing but the number of false positives increases and the
overall WER also increases.

\begin{figure*}
\begin{centering}
\includegraphics[width=1\textwidth]{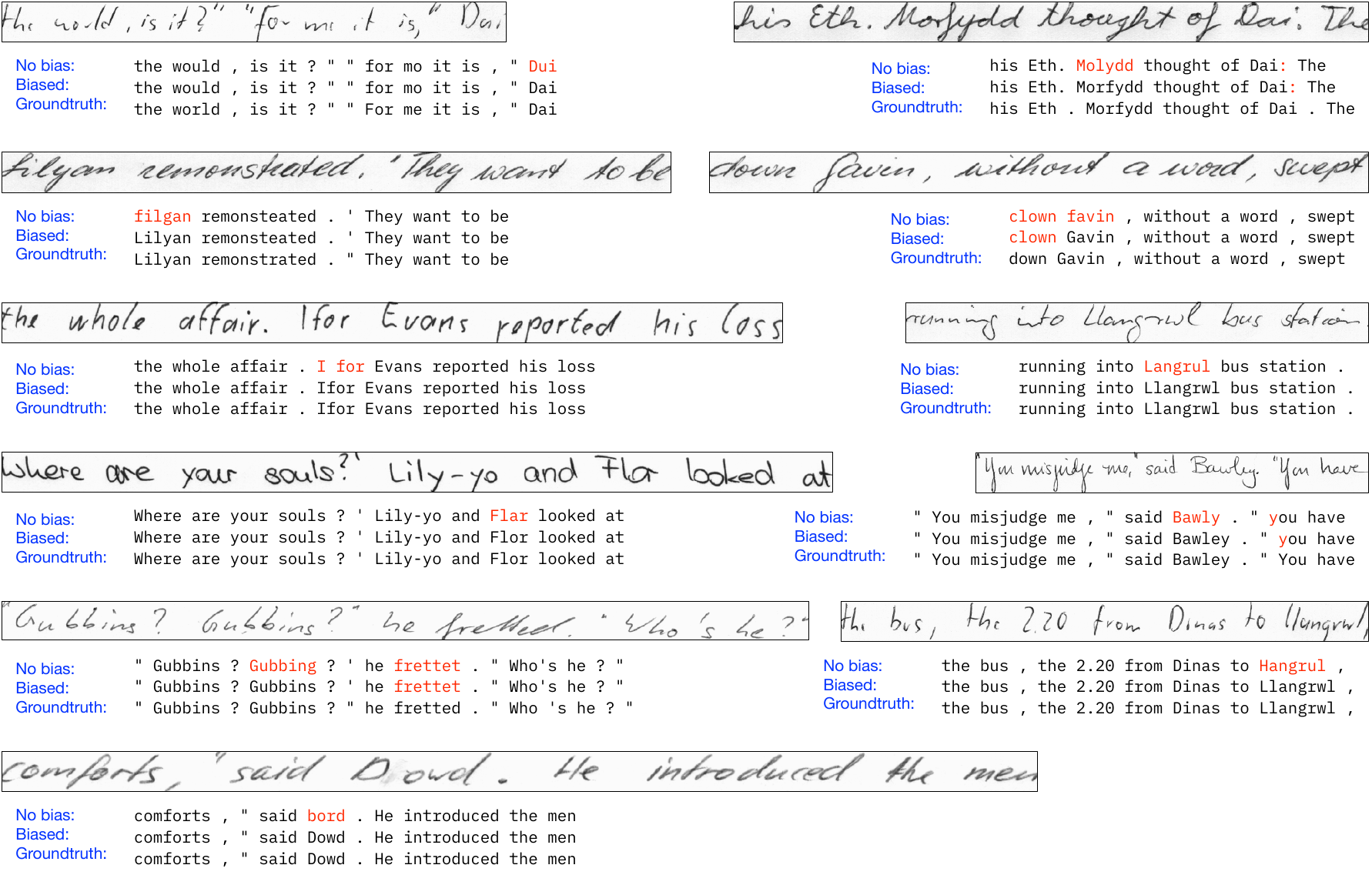}
\par\end{centering}
\caption{Examples of regex biasing on IAM. ``Biased'' results are from regex
biasing with $\alpha=-5$. Recognition errors are highlighted in red
color. Images are framed by thin black lines for clarity.}

\label{fig:iam-examples}
\end{figure*}

Some examples are shown in Figure~\ref{fig:iam-examples}. With regex
biasing, the recognizer can recognize highly ambiguous words from
the domain vocabulary, while not affecting the recognition of other
words.

\begin{table*}
\caption{Name error rate (\%) and full set WER (\%) on IAM.}

\medskip{}

\begin{centering}
\begin{tabular}{|c||c|c|c|c|c|c|c|}
\hline 
{\footnotesize{}Subset} & {\footnotesize{}no bias} & {\footnotesize{}$\alpha=0$} & {\footnotesize{}$-1$} & {\footnotesize{}$-3$} & {\footnotesize{}$-5$} & {\footnotesize{}$-10$} & {\footnotesize{}$-20$}\tabularnewline
\hline 
\hline 
\multirow{2}{*}{{\footnotesize{}Names}} & \multirow{2}{*}{{\footnotesize{}36.0}} & {\footnotesize{}28.0} & {\footnotesize{}23.7} & {\footnotesize{}18.0} & {\footnotesize{}15.7} & {\footnotesize{}11.7} & {\footnotesize{}10.3}\tabularnewline
\cline{3-8} \cline{4-8} \cline{5-8} \cline{6-8} \cline{7-8} \cline{8-8} 
 &  & \textcolor{blue}{\footnotesize{}-22\%} & \textcolor{blue}{\footnotesize{}-34\%} & \textcolor{blue}{\footnotesize{}-50\%} & \textcolor{blue}{\footnotesize{}-56\%} & \textcolor{blue}{\footnotesize{}-68\%} & \textcolor{blue}{\footnotesize{}-71\%}\tabularnewline
\hline 
\multirow{2}{*}{{\footnotesize{}Full}} & \multirow{2}{*}{{\footnotesize{}14.3}} & {\footnotesize{}14.2} & {\footnotesize{}14.1} & {\footnotesize{}14.0} & {\footnotesize{}14.0} & {\footnotesize{}14.0} & {\footnotesize{}15.8}\tabularnewline
\cline{3-8} \cline{4-8} \cline{5-8} \cline{6-8} \cline{7-8} \cline{8-8} 
 &  & \textcolor{blue}{\footnotesize{}-1\%} & \textcolor{blue}{\footnotesize{}-1\%} & \textcolor{blue}{\footnotesize{}-2\%} & \textcolor{blue}{\footnotesize{}-2\%} & \textcolor{blue}{\footnotesize{}-3\%} & \textcolor{red}{\footnotesize{}+10\%}\tabularnewline
\hline 
\end{tabular}
\par\end{centering}
\label{tab:results-iam}
\end{table*}

\subsection{Runtime analysis}

\begin{table}
\caption{Runtime analysis of regex biasing on different datasets. Numbers are
the average time (in milliseconds) for recognizing one text line.}

\medskip{}

\begin{centering}
\begin{tabular}{|c||c|c|c|c|}
\hline 
 & {\footnotesize{}no bias} & {\footnotesize{}$\alpha=0$} & {\footnotesize{}$-1$} & {\footnotesize{}$-5$}\tabularnewline
\hline 
\hline 
{\footnotesize{}Driver License} & {\footnotesize{}1.6} & {\footnotesize{}1.6} & {\footnotesize{}1.6} & {\footnotesize{}1.5}\tabularnewline
\hline 
{\footnotesize{}Passport MRZ} & {\footnotesize{}14.6} & {\footnotesize{}12.2} & {\footnotesize{}5.9} & {\footnotesize{}1.0}\tabularnewline
\hline 
{\footnotesize{}IAM} & {\footnotesize{}6.8} & {\footnotesize{}7.1} & {\footnotesize{}6.8} & {\footnotesize{}6.8}\tabularnewline
\hline 
\end{tabular}
\par\end{centering}
\label{tab:runtime-analysis}
\end{table}

Finally, we analyze the impact of regex biasing on inference time
in Table~\ref{tab:runtime-analysis}. Overall, the amount of added
time by regex biasing ranges from less than 1ms to negative. On the
driver license dataset, regex biasing has little impact on runtime,
and as $\alpha$ decreases, the inference time decreases. This is
because a lower $\alpha$ reduces the number of search paths and therefore
accelerates the search process. This phenomenon is more pronounced
on the passport dataset, where the inference time drops by 10 times
as $\alpha$ lowers.

\section{Conclusion}

We have proposed a novel method for biasing a recognizer using regular
expressions. This method improves the performance of a recognizer
on domain-specific data with efficacy, requires no labeled data and
training process, and has limited impact on runtime speed.

With some modifications, the decoder we have used in this paper may
also work with auto-regressive text recognition models, such as attention-based
recognizers~\cite{LitmanATLMM20,ShiYWLYB19,QiaoZYZ020} and RNN transducer
models~\cite{abs-1211-3711}. This can be explored in the future.

Structured text comes in many other forms, where the proposed biasing
method may find its usage. For example, math equations (represented
by Latex code) are strongly structured. A recognizer may benefit from
limiting its search space to the one defined by the Latex syntax.
We are also interested in exploring this direction in the future.

\bibliographystyle{ieee_fullname}
\bibliography{references}

\begin{thebibliography}{10}\itemsep=-1pt

\bibitem{AhoSU86}
Alfred~V. Aho, Ravi Sethi, and Jeffrey~D. Ullman.
\newblock {\em Compilers: Principles, Techniques, and Tools}.
\newblock Addison-Wesley series in computer science / World student series
  edition. Addison-Wesley, 1986.

\bibitem{AleksicAEKCM15}
Petar~S. Aleksic, Cyril Allauzen, David Elson, Aleksandar Kracun, Diego~Melendo
  Casado, and Pedro~J. Moreno.
\newblock Improved recognition of contact names in voice commands.
\newblock In {\em {ICASSP} 2015}, pages 5172--5175. {IEEE}, 2015.

\bibitem{AllauzenRSSM07}
Cyril Allauzen, Michael Riley, Johan Schalkwyk, Wojciech Skut, and Mehryar
  Mohri.
\newblock Openfst: {A} general and efficient weighted finite-state transducer
  library.
\newblock In Jan Holub and Jan Zd{\'{a}}rek, editors, {\em {CIAA} 2007}, volume
  4783 of {\em Lecture Notes in Computer Science}, pages 11--23. Springer,
  2007.

\bibitem{AroraGWMSKCRBPE19}
Ashish Arora, Paola Garc{\'{\i}}a, Shinji Watanabe, Vimal Manohar, Yiwen Shao,
  Sanjeev Khudanpur, Chun{-}Chieh Chang, Babak Rekabdar, Bagher BabaAli, Daniel
  Povey, David Etter, Desh Raj, Hossein Hadian, and Jan Trmal.
\newblock Using {ASR} methods for {OCR}.
\newblock In {\em {ICDAR} 2019}, pages 663--668. {IEEE}, 2019.

\bibitem{BrownPdLM92}
Peter~F. Brown, Vincent J.~Della Pietra, Peter~V. de Souza, Jennifer~C. Lai,
  and Robert~L. Mercer.
\newblock Class-based n-gram models of natural language.
\newblock {\em Comput. Linguistics}, 18(4):467--479, 1992.

\bibitem{CaiH17}
Meng Cai and Qiang Huo.
\newblock Compact and efficient wfst-based decoders for handwriting
  recognition.
\newblock In {\em {ICDAR} 2017}, pages 143--148. {IEEE}, 2017.

\bibitem{pykaldi}
Dogan Can, Victor~R. Martinez, Pavlos Papadopoulos, and Shrikanth~S. Narayanan.
\newblock Pykaldi: A python wrapper for kaldi.
\newblock In {\em {ICASSP} 2018}. IEEE, 2018.

\bibitem{0001TDCSLH17}
Kai Chen, Li Tian, Haisong Ding, Meng Cai, Lei Sun, Sen Liang, and Qiang Huo.
\newblock A compact {CNN-DBLSTM} based character model for online handwritten
  chinese text recognition.
\newblock In {\em {ICDAR} 2017}, pages 1068--1073. {IEEE}, 2017.

\bibitem{ChengBXZPZ17}
Zhanzhan Cheng, Fan Bai, Yunlu Xu, Gang Zheng, Shiliang Pu, and Shuigeng Zhou.
\newblock Focusing attention: Towards accurate text recognition in natural
  images.
\newblock In {\em {IEEE} International Conference on Computer Vision, {ICCV}
  2017, Venice, Italy, October 22-29, 2017}, pages 5086--5094. {IEEE} Computer
  Society, 2017.

\bibitem{chomsky2002syntactic}
Noam Chomsky.
\newblock {\em Syntactic structures}.
\newblock Walter de Gruyter, 2002.

\bibitem{CongHHG19}
Fu'ze Cong, Wenping Hu, Qiang Huo, and Li Guo.
\newblock A comparative study of attention-based encoder-decoder approaches to
  natural scene text recognition.
\newblock In {\em {ICDAR} 2019}, pages 916--921. {IEEE}, 2019.

\bibitem{abs-1211-3711}
Alex Graves.
\newblock Sequence transduction with recurrent neural networks.
\newblock {\em CoRR}, abs/1211.3711, 2012.

\bibitem{GravesFGS06}
Alex Graves, Santiago Fern{\'{a}}ndez, Faustino~J. Gomez, and J{\"{u}}rgen
  Schmidhuber.
\newblock Connectionist temporal classification: labelling unsegmented sequence
  data with recurrent neural networks.
\newblock In William~W. Cohen and Andrew~W. Moore, editors, {\em {ICML} 2006},
  volume 148 of {\em {ACM} International Conference Proceeding Series}, pages
  369--376. {ACM}, 2006.

\bibitem{HaynorA20}
Ben Haynor and Petar~S. Aleksic.
\newblock Incorporating written domain numeric grammars into end-to-end
  contextual speech recognition systems for improved recognition of numeric
  sequences.
\newblock In {\em {ICASSP} 2020}, pages 7809--7813. {IEEE}, 2020.

\bibitem{KudoR18}
Taku Kudo and John Richardson.
\newblock Sentencepiece: {A} simple and language independent subword tokenizer
  and detokenizer for neural text processing.
\newblock In Eduardo Blanco and Wei Lu, editors, {\em {EMNLP} 2018}, pages
  66--71. Association for Computational Linguistics, 2018.

\bibitem{LiaoZWXLLYB19}
Minghui Liao, Jian Zhang, Zhaoyi Wan, Fengming Xie, Jiajun Liang, Pengyuan Lyu,
  Cong Yao, and Xiang Bai.
\newblock Scene text recognition from two-dimensional perspective.
\newblock In {\em {AAAI} 2019}, pages 8714--8721. {AAAI} Press, 2019.

\bibitem{LitmanATLMM20}
Ron Litman, Oron Anschel, Shahar Tsiper, Roee Litman, Shai Mazor, and R.
  Manmatha.
\newblock {SCATTER:} selective context attentional scene text recognizer.
\newblock In {\em {CVPR} 2020}, pages 11959--11969. {IEEE}, 2020.

\bibitem{MartiB02}
Urs{-}Viktor Marti and Horst Bunke.
\newblock The iam-database: an english sentence database for offline
  handwriting recognition.
\newblock {\em Int. J. Document Anal. Recognit.}, 5(1):39--46, 2002.

\bibitem{MohriPR02}
Mehryar Mohri, Fernando Pereira, and Michael Riley.
\newblock Weighted finite-state transducers in speech recognition.
\newblock {\em Comput. Speech Lang.}, 16(1):69--88, 2002.

\bibitem{povey2011kaldi}
Daniel Povey, Arnab Ghoshal, Gilles Boulianne, Lukas Burget, Ondrej Glembek,
  Nagendra Goel, Mirko Hannemann, Petr Motlicek, Yanmin Qian, Petr Schwarz,
  et~al.
\newblock The kaldi speech recognition toolkit.
\newblock In {\em IEEE 2011 workshop on automatic speech recognition and
  understanding}, number CONF. IEEE Signal Processing Society, 2011.

\bibitem{QiaoZYZ020}
Zhi Qiao, Yu Zhou, Dongbao Yang, Yucan Zhou, and Weiping Wang.
\newblock {SEED:} semantics enhanced encoder-decoder framework for scene text
  recognition.
\newblock In {\em {CVPR} 2020}, pages 13525--13534. {IEEE}, 2020.

\bibitem{RoarkSARST12}
Brian Roark, Richard Sproat, Cyril Allauzen, Michael Riley, Jeffrey Sorensen,
  and Terry Tai.
\newblock The opengrm open-source finite-state grammar software libraries.
\newblock In {\em {ACL} 2012}, pages 61--66. The Association for Computer
  Linguistics, 2012.

\bibitem{SennrichHB16a}
Rico Sennrich, Barry Haddow, and Alexandra Birch.
\newblock Neural machine translation of rare words with subword units.
\newblock In {\em {ACL} 2016}. The Association for Computer Linguistics, 2016.

\bibitem{ShiBY17}
Baoguang Shi, Xiang Bai, and Cong Yao.
\newblock An end-to-end trainable neural network for image-based sequence
  recognition and its application to scene text recognition.
\newblock {\em {IEEE} Trans. Pattern Anal. Mach. Intell.}, 39(11):2298--2304,
  2017.

\bibitem{ShiYWLYB19}
Baoguang Shi, Mingkun Yang, Xinggang Wang, Pengyuan Lyu, Cong Yao, and Xiang
  Bai.
\newblock {ASTER:} an attentional scene text recognizer with flexible
  rectification.
\newblock {\em {IEEE} Trans. Pattern Anal. Mach. Intell.}, 41(9):2035--2048,
  2019.

\end{thebibliography}

\end{document}